\documentclass[10pt,twocolumn,letterpaper]{article}

\usepackage{iccv}
\usepackage{times}
\usepackage{epsfig}
\usepackage{graphicx}
\usepackage{amsmath}
\usepackage{amssymb}

\usepackage[breaklinks=true,bookmarks=false]{hyperref}

\usepackage{cite}
\usepackage{graphicx}
\usepackage{amsmath,amssymb} 
\usepackage{color}
\usepackage{times}
\usepackage{epsfig}
\usepackage{algorithm}
\usepackage{algorithmic}
\usepackage{caption}
\usepackage{epstopdf}
\usepackage{hyperref}
\hypersetup{
colorlinks=true,
urlcolor=blue,
}
\DeclareGraphicsExtensions{.pdf,.png,.jpg}
\usepackage{ragged2e}
\usepackage{makecell}
\usepackage[normalem]{ulem}
\usepackage{amsmath}
\usepackage{bbm}
\newcommand{\stkout}[1]{{\textcolor{red}\ifmmode\text{\sout{\ensuremath{#1}}}\else\sout{#1}\fi}}\usepackage{times}
\usepackage{epsfig}
\usepackage{graphicx}
\usepackage{amsmath}
\usepackage{amssymb}
\usepackage{multirow}
\usepackage{tabularx}
\usepackage[table,xcdraw]{xcolor}
\usepackage{tablefootnote}

\newcommand{\OURS}{OLN}

\newcommand{\figureref}[1]{Figure~\ref{#1}}
\newcommand{\tableref}[1]{Table~\ref{#1}}
\newcommand{\sectionref}[1]{Section~\ref{#1}}
\newcommand{\secref}[1]{§\ref{#1}}

\renewcommand{\paragraph}[1]{\vspace{1mm}\noindent\textbf{#1}}

\DeclareMathAlphabet{\pazocal}{OMS}{zplm}{m}{n}

\usepackage{amsthm}

\usepackage[switch]{lineno}
\usepackage{multirow}

\iccvfinalcopy 

\def\iccvPaperID{****} 

\ificcvfinal\fi

\begin{document}

\title{Learning Open-World Object Proposals without Learning to Classify}

\author{Dahun Kim\\
KAIST\\
\and
Tsung-Yi Lin\\
Google Brain\\

\and
Anelia Angelova\\
Google Brain\\

\and
In So Kweon\\
KAIST\\

\and
Weicheng Kuo\\
Google Brain\\
}

\maketitle
\ificcvfinal\thispagestyle{empty}\fi

\begin{abstract}
Object proposals have become an integral pre-processing steps of many vision pipelines including object detection, weakly supervised detection, object discovery, tracking, etc. Compared to the learning-free methods, learning-based proposals have become popular recently due to the growing interest in object detection. The common paradigm is to learn object proposals from data labeled with a set of object regions and their corresponding categories. However, this approach often struggles with novel objects in the open world that are absent in the training set. In this paper, we identify that the problem is that the binary classifiers in existing proposal methods tend to overfit to the training categories. 
Therefore, we propose a \textbf{classification-free} Object Localization Network (OLN) which estimates the objectness of each region purely by how well the location and shape of a region overlap with any ground-truth object (e.g., centerness and IoU). This simple strategy learns generalizable objectness and outperforms existing proposals on cross-category generalization on COCO, as well as cross-dataset evaluation on RoboNet, Object365, and EpicKitchens. Finally, we demonstrate the merit of OLN for long-tail object detection on large vocabulary dataset, LVIS, where we notice clear improvement in rare and common categories.
\end{abstract}

\section{Introduction}
%
%
%
%
Object proposals are a set of regions or bounding boxes that contain objects with high likelihood \cite{kuo2015deepbox,pinheiro2015learning,pont2016multiscale,fasterNIPS2015,uijlings2013selective, vu2019cascade, wang2019region, zitnick2014edge}. They have become the integral pre-processing steps for many computer vision systems, including object detection \cite{kuo2015deepbox,pinheiro2015learning,pont2016multiscale,fasterNIPS2015,uijlings2013selective, zitnick2014edge}, segmentation \cite{arbelaez2012semantic,carreira2012semantic,dai2015convolutional}, object discovery \cite{deselaers2012weakly,cho2015unsupervised,rubinstein2013unsupervised}, weakly supervised object detection \cite{arun2019dissimilarity,gao2019c,tang2018pcl}, visual tracking \cite{kwak2015unsupervised,wang2015transferring}, content-aware retargeting \cite{sun2011scale}, etc. Due to the success of object detection, the recent trend in object proposal research has shifted from object discovery to detection. While the goal of object discovery proposals is to propose \textit{any} objects in the image, the goal of detection proposals is to propose only the labeled categories for downstream classifier. Learning-based proposals are popular detection proposals because of simplicity and shared computation with downstream detection. However, unlike their learning-free counterparts \cite{pont2016multiscale,uijlings2013selective,zitnick2014edge}, these methods tend to overfit to annotated categories and struggle with novel objects \cite{kuo2015deepbox,pinheiro2015learning,wang2020leads}. We want to ask, is it possible to combine the best of both worlds and ``\textit{learn open-world (novel) object proposals?}'' This could potentially unlock learning-based proposals for promising applications including open-world detection~\cite{joseph2021towards} / segmentation~\cite{wang2021unidentified}, robot grasping \cite{dasari2019robonet}, egocentric video understanding \cite{Damen2020Collection}, and large vocabulary detection \cite{gupta2019lvis}.

\begin{figure}
\includegraphics[width=\linewidth]{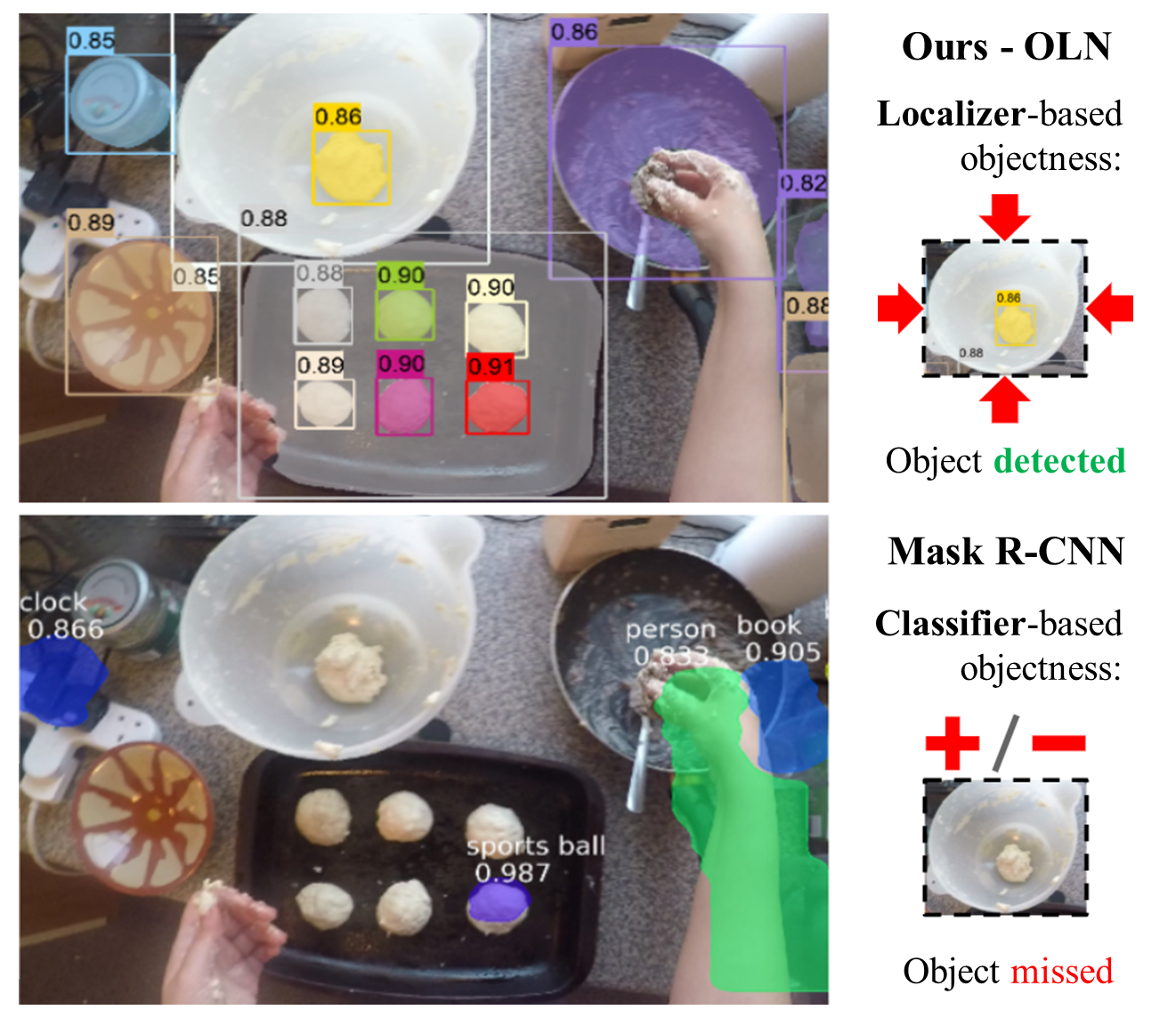}
\caption{\small{Humans can recognize novel objects in this image despite having never seen them before. ``Is it possible to learn open-world (novel) object proposals?'' In this paper we propose Object Localization Network (OLN) that learns localization cues instead of foreground vs background classification. Only trained on COCO\cite{lin2014microsoft}, \OURS{} is able to propose many novel objects (top) missed by Mask R-CNN (bottom) on an out-of-sample frame in an ego-centric video.
}}
\label{fig:teaser}
\end{figure}

Given a set of object annotations, we want to learn what general objects look like and propose highly dissimilar object candidates from unseen categories and new data sources. This matches the ability of human to detect novel objects in new environments without naming their categories \eg a piece of obstacle on the road, a novel product on the shelf. Our main insight is that the classifiers in existing object proposers \cite{kuo2015deepbox,pinheiro2015learning,vu2019cascade} or class agnostic detectors \cite{fasterNIPS2015,tian2019fcos} impedes such generalization, because the model tends to overfit to labeled objects and treat the unlabeled objects in the training set as background. We propose Object Localization Network (OLN), which learns to detect objects by predicting how well a region is localized instead of performing foreground-background classification. This simple idea allows the model to learn stronger objectness cues. To our best knowledge, we are the first to demonstrate the value of learning \textit{pure localization-based objectness for proposing novel objects}, although the idea of incorporating the localization quality estimation has been proposed by others in the standard fixed-category detection setting \cite{huang2019mask,jiang2018acquisition,tian2019fcos,wu2020iou}. We show that a \textit{classifier-free} object proposer is  key to achieve optimal cross-category and cross-dataset generalization, which is an important design difference to existing proposers or class agnostic detectors.

We study the efficacy of \OURS{} on the COCO cross category setting following existing works \cite{kuo2015deepbox,pinheiro2015learning,wang2020leads}. Despite the simplicity, OLN outperforms the state-of-the-art by \textbf{+3.3 AUC} (+5.0 AR@10, +5.1 AR@100) on novel categories. Our ablation studies confirm that the use of foreground-vs-background classifier hurts, and that localization helps. In addition, we study cross-dataset generalization from COCO to RoboNet~\cite{dasari2019robonet}, Objects365~\cite{shao2019objects365}, and EpicKitchens~\cite{Damen2020Collection}. We chose RoboNet because it contains a wide range of novel objects common in robotics grasping application and the bin environment permits more reliable exhaustive annotation for proper evaluation \cite{chavali2016object}. On RoboNet, \OURS{} performs exhaustive, class-agnostic object detection and outperforms the standard approach by \textbf{+13$\sim$16 AP}, whereas on Objects365 \OURS{} has +4 AR@10 and +8 AR@100 over the standard approach. Qualitative visualization on EpicKitchens \cite{Damen2020Collection} further shows that \OURS{} outperforms the standard approach in detecting a variety of novel objects. Last but not least, we apply \OURS{} as a drop-in replacement for RPN\cite{fasterNIPS2015} on LVIS long tail detection \cite{gupta2019lvis} and observe a gain of +1.4 AP, mostly attributed to the rare (\textbf{+3.4 APr}) and common categories (+1.8 APc). This shows \OURS{} is able to capture the long tail in large vocabulary detection. 

It is worth noting that estimating localization quality  is  not  new  in  the  standard  detection,  but  they  are always used alongside classification and validated on seen categories only, e.g.  FCOS~\cite{tian2019fcos}. To our knowledge, we are the first to explore the use of localization cues independent of classification for object proposals. This discovery helps us obtain notable gains on COCO and generalize to many dissimilar datasets better than existing method.

Our contributions can be summarized as follows: 
\begin{itemize}
\item To our knowledge, we are the first to show the value of pure localization-based objectness learning for novel object proposals, and propose a simple-yet-effective classifier-free Object Localization Network (OLN).
\item Our approach outperforms state-of-the-art methods on cross-category setting on COCO and improves cross-dataset settings on RoboNet and Object365, long-tail detection (LVIS) and egocentric videos (EpicKitchens) over the standard approach.
\item We carefully annotated the RoboNet dataset for the presence of all objects in an exhaustive  fashion. We perform open-world class-agnostic object detection, and evaluate the Average Precision, which also improves existing AR-based evaluation of proposals on partially-annotated data.
\item Extensive ablation and analysis on OLN modeling choices reveal the benefits of each localization cue and the overfitting of existing classifier-based methods.
\end{itemize}

\section{Related Work}
Below we discuss existing efforts to improve proposal and detection quality, and works that scale detection to more visual categories.

\noindent\textbf{Object proposal.}\quad
In early works, the emphasis was on category-independent object proposals~\cite{alexe2012measuring,arbelaez2014multiscale,endres2013category,manen2013prime, zitnick2014edge}, where the goal is to identify instances of all objects in the image irrespective of their category. These works utilize hand-crafted heuristics to capture the notion of general objects~\cite{uijlings2013selective, zitnick2014edge}, \ie, color contrast, edge. 

Recently, learning-based proposals~\cite{gidaris2016attend,kuo2015deepbox,li2019zoom, pinheiro2015learning,fasterNIPS2015} have demonstrated better performance than classical approaches in both precision and recall, and are an important part of two-stage detectors. A representative example is region proposal network (RPN)~\cite{fasterNIPS2015} which identifies a set of regions in a given image that could contain objects, which are then used by a downstream detector module to localize and classify objects. A number of follow-up works~\cite{gidaris2016attend,vu2019cascade,wang2019region,yang2016craft} have been proposed to improve the quality of such region proposals and reduce their number in order to speed up the final detection task. In fact, these proposal modules are trained end-to-end with the detector module, where the notion of objectness is defined by a set of training categories in the dataset. Despite their progress in detecting objects of the known, supervised categories, learning-based proposals still struggle on novel objects. 

More closely related to our work are studies on generalization of object proposals onto unseen classes. Chavali et al.~\cite{chavali2016object} demonstrate the standard proposal evaluation is problematic or ``gameable'' in evaluating category-independence of object proposal because detection of unknown object classes is explicitly penalized in the benchmark protocol. Wang et al.~\cite{wang2020leads} study the generalization from a dataset standpoint and demonstrate the impact of visual diversity and label granularity of training dataset on the generalization of object proposers. In contrast, we focus on the modeling choices in designing an object proposer that can generalize to novel categories and new datasets.

\vspace{1mm}
\noindent \textbf{Multi-class detection.} \quad
A number of efforts have been made to scale up the number of classes for detection by transferring commonalities between object categories with varying degrees and quality of supervision. 

Weakly-supervised approaches~\cite{arun2019dissimilarity,bilen2016weakly,diba2017weakly,tang2017multiple,wang2018collaborative} aim to utilize abundant image-level labels and leverage class-agnostic box proposals to build detectors. Approaches under semi-supervised setup~\cite{hoffman2014lsda,redmon2017yolo9000,singh2018r,singh2018dock,tang2016large,uijlings2018revisiting,yang2019detecting} employ the weak image-level labels for novel classes as well as box-level labels for base classes. For example, YOLO-9000~\cite{redmon2017yolo9000} and R-FCN-3000~\cite{singh2018r} concurrently train on box-level and image-level data to scale up the detector’s class coverage. Knowledge transfer-based methods learn to transfer the proposals from base to novel classes based on their similarity on semantic hierarchy. This line of research is also related to few-shot~\cite{chen2018lstd,kang2019few,wang2020frustratingly,wang2019meta,yan2019meta} and zero-shot detection\cite{bansal2018zero,frome2013devise,rahman2018zero,xian2018zero} methods, which attempt to detect novel classes given only a few samples or class descriptions.

In contrast to the above class-specific detectors, our goal is to go beyond the concept of category and detect all objects in a category-independent manner (without classification). Even though the multi-class detectors enumerate many categories, they still fail to generalize for unseen/unknown object categories. 

\paragraph{Dataset efforts for detection at scale.} \quad
Existing datasets focus on single-dataset settings~\cite{Everingham15, gupta2019lvis, kuznetsova2018open,lin2014microsoft, shao2019objects365}.
Models trained on one dataset are only evaluated on the same dataset. Recently the Robust Vision Challenge~\cite{rvc2020} is a first step towards cross-dataset benchmark. We hypothesize that learning objectness on one dataset should also transfer to another dataset, as the saliency cues tend to be more generalizable than class-specific information. Therefore, in this work we study the generalization setting of training on COCO, then testing on other datasets: RoboNet~\cite{dasari2019robonet}, Objects365~\cite{shao2019objects365}, EpicKitchens~\cite{Damen2020Collection}, and LVIS~\cite{gupta2019lvis}.

\paragraph{Localization cues for object detection.} \quad
Researchers have studied many ways to improve the localization quality in object detections by learning centerness \cite{tian2019fcos}, iterative proposal refinement \cite{vu2019cascade}, or box/mask IoU prediction \cite{huang2019mask,jiang2018acquisition,tychsen2018improving}. These approaches have demonstrated meaningful improvement in the standard object detection task. However, it remains an open question whether these methods can transfer to novel categories. In addition, most of these works use these localization cues alongside classifier outputs. It is unclear whether these cues \textit{on their own} can effectively distinguish objects from background.

\section{Proposed Method}

\subsection{Baselines}

Before we describe \OURS{} in section \ref{sec:localization_objectness} and \ref{sec:oln_method}, we would like to define the baselines which can address the same unseen category generalization problem. Region proposal network (RPN)~\cite{fasterNIPS2015,vu2019cascade, wang2019region} is the most common family of approaches of objectness learning in object detection. By design, RPNs aim to propose all objects in the image regardless of their categories, but in practice they often struggle when encountered with novel objects in the open world. Another family of baseline is to train existing object detectors in a class-agnostic fashion by treating all annotated categories as one foreground category. As OLN is built upon RPN and Faster R-CNN, we use both as strong baselines throughout the paper. Moreover, we provide comparisons with different state-of-the-art models in region proposal~\cite{fasterNIPS2015,vu2019cascade,wang2019region} and object detection~\cite{he2017mask,fasterNIPS2015,tian2019fcos}. As seen in our experiments later, OLN outperforms all of them in various generalization scenarios. 

\begin{figure}
\centering
\includegraphics[width=\linewidth]{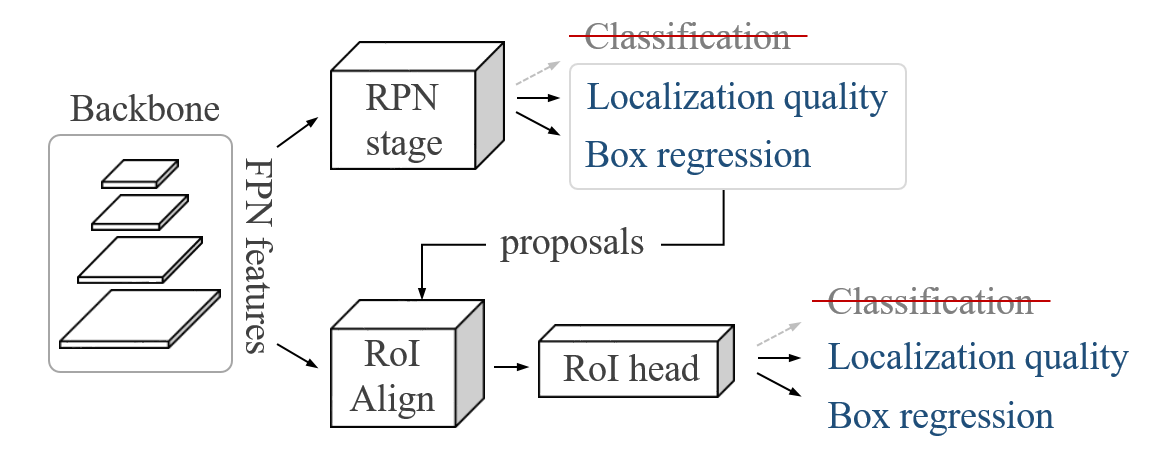}
\caption{\small{\textbf{Overview of Object Localization Network (\OURS{}).} We replace the classifier head which is common in most object proposal methods with localization quality estimators, e.g. centerness and IoU. This applies to the anchors at whole-image level and to each region of interest (RoI).}}
\label{fig:model}
\end{figure}

\subsection{Pure localization-based objectness}
\label{sec:localization_objectness}
In the context of learning-based object proposals, ``object” is defined as a set of annotated categories, and the learning of objectness is cast as a binary classification task: whether or not a region belongs to the union of the pre-defined categories. However, our main insight is that such discriminative learning of the foreground-vs-background problem impedes generalization because the model learns to classify the unlabeled/unknown objects as background. To address this problem, we propose a non-discriminative and classification-free notion of objectness.

The classification view of ``objectness'' is to ask ``how much does this region \textit{look} like a foreground object?'' From a localization standpoint, we want to ask instead ``how well does this region overlap with any ground-truth object?''. Our intuition is that every object can be characterized by its location and shape, regardless of its category. OLN leverages these geometric cues to capture the objectness of a proposed region. We demonstrate that the learnt objectness cues based on the localization (location and shape) quality can collectively improve the generalization of object proposals beyond the labeled categories and data sources. We adopt centerness~\cite{tian2019fcos} and IoU score~\cite{jiang2018acquisition} for location and shape quality measures respectively, while not restricting other choices such as Dice coefficient~\cite{milletari2016v} and generalized-IoU~\cite{rezatofighi2019generalized}.

The idea of incorporating localization quality is not totally new in object detection. Several works~\cite{huang2019mask,jiang2018acquisition,tian2019fcos,tychsen2018improving} recalibrate the final detection confidence by using both localization and classification subnets. Note that, however, their localization cues are thoroughly an \textit{auxiliary to the classifiers} and are devised for within-category detection. In contrast, we demonstrate that a \textit{pure localization-based objectness} is the key to generalize beyond category and across datasets, and that a classification head severely hurts generalization. To best our knowledge, this intuition has not been discussed in any of the prior works.

\subsection{Object Localization Network (OLN)}
\label{sec:oln_method}
The goal of OLN is to learn localization for objects and enable better generalization to new and unseen categories.
\OURS{} is a two-stage object proposer (see \figureref{fig:model}). Similar to Faster R-CNN \cite{fasterNIPS2015}, \OURS{} consists of a fully-convolutional FCN stage and region-based RoI stage, but the key difference is that the classifiers in both FPN and ROI stages are replaced with localization quality predictions. %

\begin{figure*}[]
\begin{center}
\resizebox{\linewidth}{!}{
\begin{tabular}{@{}c@{\hskip 0.002\linewidth}c@{\hskip 0.002\linewidth}c@{\hskip 0.002\linewidth}c@{\hskip 0.002\linewidth}c@{}}
\includegraphics[width=0.263\linewidth]{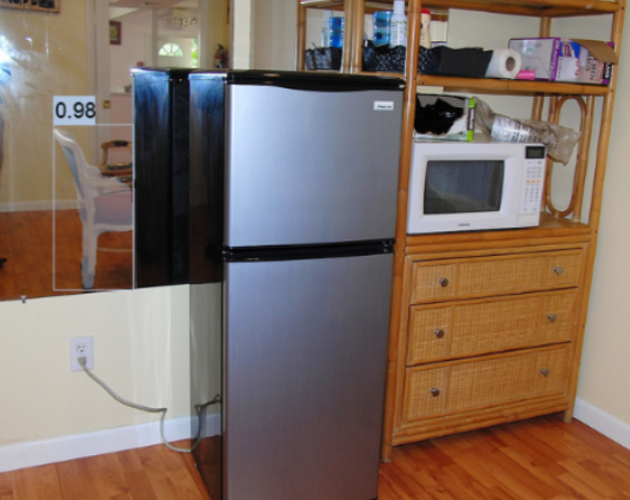} &
\includegraphics[width=0.158\linewidth]{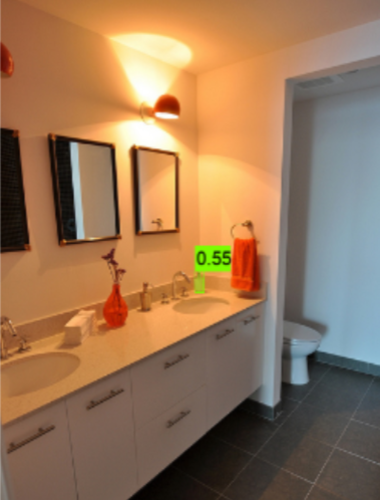} &
\includegraphics[width=0.137\linewidth]{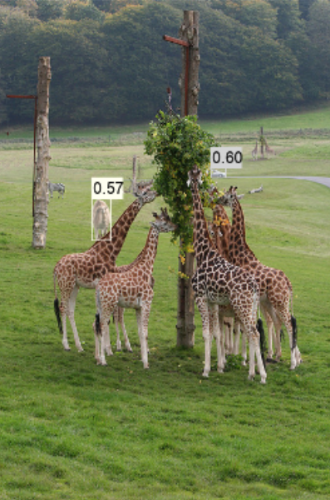} &
\includegraphics[width=0.158\linewidth]{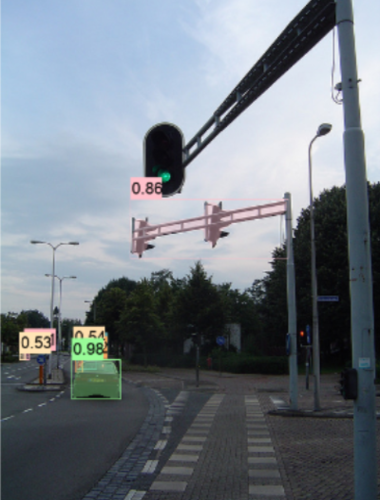} &
\includegraphics[width=0.276\linewidth]{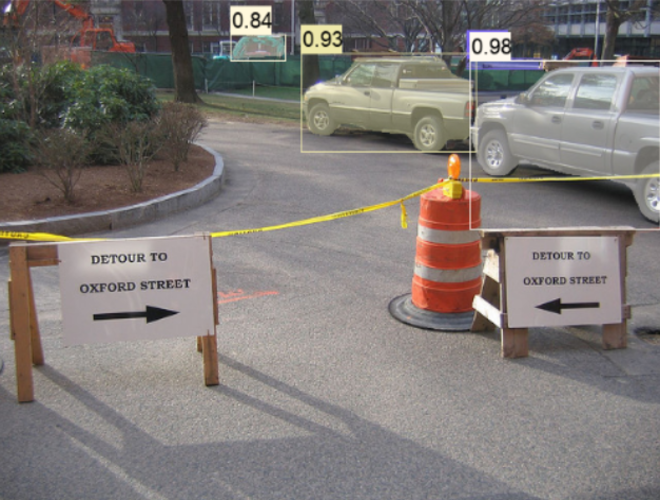}
 \\
\includegraphics[width=0.263\linewidth]{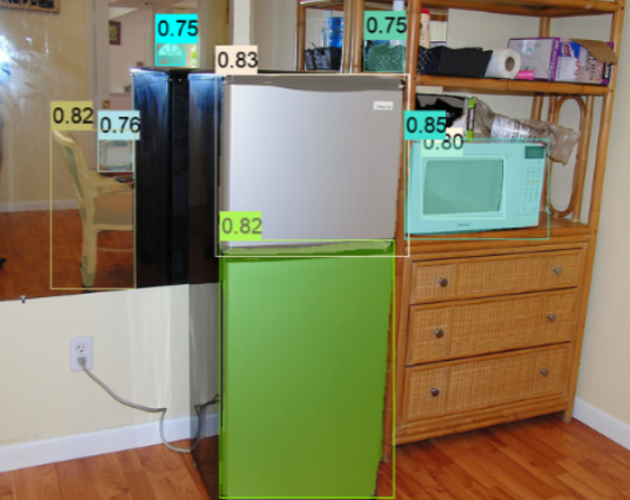} &
\includegraphics[width=0.158\linewidth]{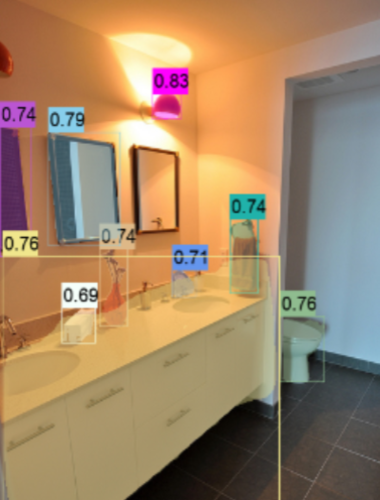} &
\includegraphics[width=0.137\linewidth]{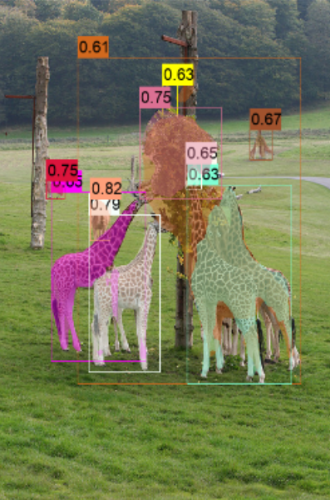} &
\includegraphics[width=0.158\linewidth]{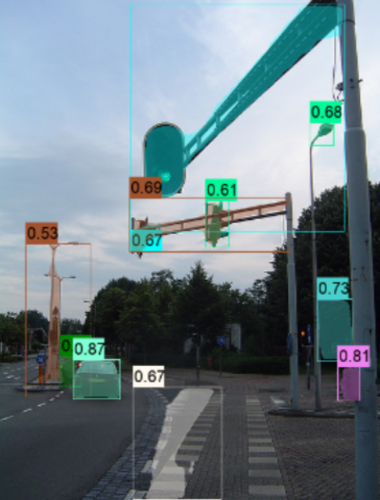} &
\includegraphics[width=0.276\linewidth]{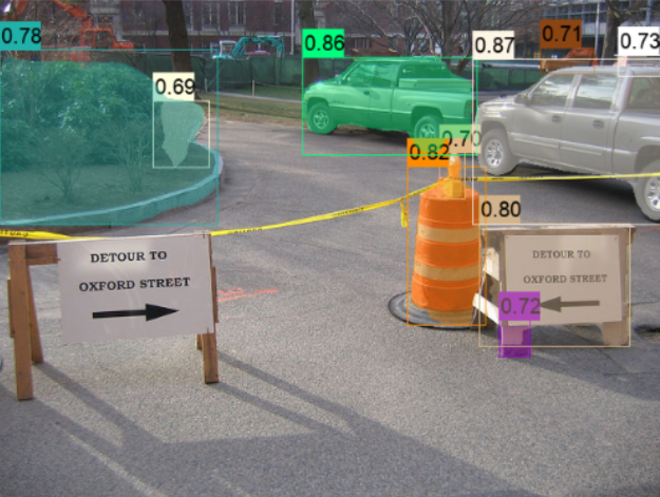}
 \\
\end{tabular}}
\vspace{-3mm}
\end{center}
\caption{\small{Visualization of \textit{VOC} $\rightarrow$ \textit{Non-VOC} category generalization. (Top: Mask R-CNN, Bottom: OLN-Mask).}}
\label{fig:coco}
\end{figure*}

\paragraph{OLN-RPN.} \quad
The input to this region proposal stage is the features from each level of ResNet feature pyramid \cite{lin2017feature}. Each feature map goes through a convolution layer followed by two separate layers, one for bounding box regression and the other for localization quality prediction. The network architecture design follows from the standard RPN heads.

We choose centerness \cite{tian2019fcos} as the localization quality target and train both heads with L1 losses. Learning localization instead of classification at the proposal stage is crucial as it avoids overfitting to the foreground by classification. For training the localization quality estimation branch, we randomly sample 256 anchors having an IoU larger than 0.3 with the matched ground-truth boxes, without any explicit background sampling. For the box regression, we replace the standard box-delta targets (\textit{xyhw}) with distances from the location to four sides of the ground-truth box (\textit{lrtb}) as in~\cite{tian2019fcos}. We choose to use one anchor per feature location as opposed to 3 in RPN, because we observe its better generalization as each anchor can ingest more data.

\paragraph{OLN-Box.} \quad
We take the top-scoring (\eg, well-centered) proposals from OLN-RPN and perform RoIAlign \cite{he2017mask} to extract the region features from each feature pyramid level. Then we linearize each region features and feed it through two fc layers, followed by two separate fc layers, one for bounding box regression and the other for localization quality prediction. We use the same network architecture as Faster R-CNN heads \cite{fasterNIPS2015}. We choose IoU as the localization quality target and train both heads with L1 losses. Learning localization quality at the second stage is integral as it allows the model to refine the proposal scoring and simultaneously avoid overfitting to the foreground. Compared to IoU-Net which requires manual proposal generation for IoU training, \OURS{} directly computes the IoU targets from the OLN-RPN proposals and ground-truth boxes, thus greatly saving computation costs.

\noindent \textbf{Extension - \OURS{}-Mask.} \quad
We explore whether more localization learning can further improve the generalization ability of our framework. To this end, we extend our \OURS{}-Box model to perform mask prediction by adding the class-agnostic FCN mask head of Mask R-CNN, which we refer to as \OURS{}-Mask model. Following our philosophy of \OURS{}, and similarly to MS R-CNN~\cite{huang2019mask}, we learn to regress the IoU between the predicted and its GT mask. 

Our mask-IoU predictor directly branches out from the fourth layer of the added FCN mask head, without having a feedback connection from the mask prediction unlike~\cite{huang2019mask}. The IoU branch consists of a $3\times3$ convolution layer, a max pooling layer and three fully connected layers. During training, we assume mask annotations are available for the training categories, and use smooth-L1 loss for IoU regression.

\paragraph{Inference.}\quad 
During inference, the objectness score of a region $s$ is computed as a geometric mean of the centerness $c$ and IoU scores (box:$b$, mask:$m$) estimated by OLN-RPN and OLN-Box branches. For OLN-Box, the score $s = \sqrt{(c \cdot b)}$. For OLN-Mask, the score is $s = \sqrt[3]{(c \cdot b \cdot m)}$.

\section{Experiments}
We study the generalization ability of the learned object proposal networks in three challenging generalization scenarios. We outperform all methods and strong baselines on them, and in some cases, significantly. First, we study \textit{\textbf{1) cross-category generalization}} on COCO dataset by evaluating average recall (AR) on new unseen classes. We compare with other state-of-the-art proposal and detection methods, and provide extensive ablation and analysis on OLN design. Also, we explore more challenging \textit{\textbf{2) open-world class-agnostic detection}} where the setup requires to detect all objects in an exhaustive and class-agnostic fashion. The testing images contain highly dissimilar objects to those in the training dataset; We train a model on COCO and test on RoboNet dataset~\cite{dasari2019robonet} and evaluate the Average Precision (AP). We demonstrate more \textit{\textbf{3) cross-dataset generalization}} results by testing on Object365~\cite{shao2019objects365} and EpicKitchens~\cite{Damen2020Collection} datasets. Finally, we study the \textit{\textbf{4) impact on long-tail object detection}} of OLN on LVIS~\cite{gupta2019lvis} dataset. In the following, we provide experiment setups, evaluation protocol and results for each setting. In each \secref{sec:cross_category}, \secref{sec:robonet_detection}, \secref{sec:cross_dataset}, and \secref{sec:application}, we use the same training and testing pipelines for all competing methods, for a fair comparison. All experiments use the same ResNet-50~\cite{he2016deep} with feature pyramid~\cite{lin2017feature} backbone, and the same box regression head and training hyper-parameters as the Faster R-CNN \cite{fasterNIPS2015}, unless specified otherwise. More details are in supplementary materials.

\begin{table*}[]
\centering
\resizebox{0.8\linewidth}{!}{
\begin{tabular}{l|l|c|ccccc}
\hline
\multicolumn{2}{r|}{} & {\small{AUC}} & \small{AR10}  & \small{AR30}  & \small{AR100}  & \small{AR300} & \small{AR1k} \\
\hline \hline

{\textit{\textbf{Learning-free method}}}    & 
MCG~\cite{pont2016multiscale}                & 18.9  & 10.4   &16.6  & 24.9  & 33.4  & 43.8   \\
\hline

\multirow{4}{*}{\textit{\textbf{Single-stage models}}}    & 
RPN~\cite{fasterNIPS2015}                  & 14.6 & 7.4   & 12.4  & 20.0  & 27.3  & 32.2   \\
& RPN w/ 1-or-0 reg.~\cite{fasterNIPS2015}    & 14.6   & 7.5   & 13.1  & 19.5  & 27.2  & 32.4   \\
& FCOS~\cite{tian2019fcos}                   & 18.3 & 10.5  & 16.5  & 24.4  & 32.7  & 39.5   \\
& \textbf{OLN-RPN (1-stage)}                 & \textbf{19.8}  & \textbf{11.7}  & \textbf{19.4}  & \textbf{27.4}  & \textbf{33.7}  & \textbf{39.7} \\
\hline

\multirow{5}{*}{\textit{\textbf{Multi-stage models}}} &
Faster R-CNN~\cite{fasterNIPS2015}       & 19.4  & 11.7  & 17.4  & 25.5  & 34.1  & 42.2 \\
& Faster R-CNN {*filtered*}       & 21.3  & 13.3  & 20.2  & 28.7  & 36.9  & 44.3 \\
& Guided Anchoring RPN~\cite{wang2019region}      & 20.9    & 11.9  & 18.2  & 27.7  & 37.5  & \textbf{46.7}  \\
& Cascade RPN~\cite{vu2019cascade}          & 21.0 & 12.6  & 19.0  & 27.7  & 36.7  & 46.4 \\
& \textbf{OLN-Box (2-stage)}        & \textbf{24.2}  & \textbf{17.7}  & \textbf{26.0}  & \textbf{32.7}  & \textbf{38.1}  & 45.1  \\
\hline
\end{tabular}}
\caption{\small{AR results of {VOC} $\rightarrow$ {Non-VOC} generalization.}}
\label{tab:sota}
\end{table*}

\subsection{Cross-category generalization}
\label{sec:cross_category}

\paragraph{Setup.} \quad 
We split the COCO dataset into 20 seen (VOC) classes and 60 unseen (non-VOC) classes. We train a model with box annotations of only seen classes, and evaluate the recall on \textit{unseen} non-VOC classes only. To avoid evaluating any recall on seen-class objects, we do \textit{\textbf{not}} count those seen-class detection boxes into the budget $k$ when computing the Average Recall (AR@k) scores.

\paragraph{Comparison with other learning-based methods.} \quad 
We compare with learning-based single-stage and multi-stage methods. The comparison is in \tableref{tab:sota}. For \textit{all} methods, we use the standard official models available in MMDetection~\cite{mmdetection} and train with the same default $1\times$ training and testing pipeline. Batch size of 16 (2 per GPU) and an initial learning rate of 0.02 are used. Note that the \textit{detection} models are their \textit{class-agnostic} versions with a binary classifier. The NMS threshold is set to 0.7.

OLN-RPN improves the standard RPN~\cite{fasterNIPS2015} by a large margin of \textbf{+5.2} AUC (+4.3 AR10 and +7.4 AR100), and outperforms all single-stage competitors. We also include RPN trained with 1-or-0 linear regression L1 loss instead of cross entropy loss and show that simply changing the loss function does \textbf{not} help. FCOS~\cite{tian2019fcos} with a binary classifier is a popular example of combining classification and localization cue (centerness). Two-stage OLN-Box shows a large gain of \textbf{+4.9} AUC (+6.1 AR10 and +7.3 AR100) over Faster R-CNN. Guided anchoring (GA-RPN)~\cite{wang2019region} is an advanced iterative RPN with deformable convolution, and Cascade RPN~\cite{vu2019cascade} is the state-of-the-art RPN method with multi-stage refinement. OLN-Box outperforms all of them with a healthy margin: \textbf{+3.3} AUC. In terms of model size, OLN-RPN is the same size as RPN, and OLN-Box is the same as Faster R-CNN.

\paragraph{Comparison with stronger baselines.} \quad
We can come up with a stronger Faster R-CNN baseline where we do not suppress unseen classes in the classification. We filter out all unseen class objects to train Faster R-CNN *filtered* (BG sampling at IoU = 0 with unseen classes). Table 1 shows that OLN does better by learning to localize (+3.0 AUC), even when Faster R-CNN *filtered* has access to the ground truth boxes of novel objects.

It's worth noting that this paper's focus is on objectness learning for generalization of proposal models; we do not explore orthogonal factors such as advanced convolutions for feature alignment~\cite{ vu2019cascade,wang2019region} and the use of box regression statistics~\cite{vu2019cascade}, that may further improve performance.

\begin{table}[t]
\centering
\resizebox{0.96\linewidth}{!}{
\begin{tabular}{l|cccc}
\hline
                                         & 
                                         {AR10} & 
                                         {AR100} &
                                         {AR1000} & 
                                         {AUC} \\
\hline \hline
EdgeBoxes~\cite{zitnick2014edge}        & 7.4   & 17.8  & 33.8  & 13.9 \\
Geodesic~\cite{krahenbuhl2014geodesic}  & 4.0   & 18.0  & 35.9  & 12.6 \\
Sel.Search~\cite{uijlings2013selective} & 5.2   & 16.3  & 35.7  & 12.6 \\
MCG~\cite{arbelaez2014multiscale}       & 10.1  & 24.6  & 39.8  & 18.0 \\
\hline
DeepMask20~\cite{pinheiro2015learning}\tablefootnote{We acknowledge that Pinheiro et al. use shallower network backbone (VGG16) than ours (ResNet-50). However, we show their reported numbers to provide a reference level. Both DeepMask20 and OLN-Box are trained on 20 VOC classes, and tested on all 80 classes.}  & 13.9  & 28.6  & 43.1  & 21.7 \\
\hline 
\textbf{OLN-Box}  & \textbf{27.7}  & \textbf{46.1}  & \textbf{55.7}  & \textbf{34.3} \\

\hline
\end{tabular}}
\caption{\small{Comparison with learning-free methods on All Categories}. We report average recall (AR) and AUC of learning free methods and ours vs DeepMask on {VOC} $\rightarrow$ {All} generalization. The scores of competing methods are taken from~\cite{pinheiro2015learning}, which test object proposal methods on all 80 COCO classes.} 
\label{tab:free}
\end{table}

\paragraph{Comparison with learning-free methods.} \quad 
We borrowed the DeepMask~\cite{pinheiro2015learning}
setting that evaluates on `all' categories in \tableref{tab:free}. To make a cleaner comparison, we report in \tableref{tab:sota} the MCG~\cite{pont2016multiscale} (SOTA learning-free baseline) on `unseen' split and find that OLN still maintains a healthy margin above it.

\paragraph{{Ablation: modeling choices.}}\quad
We study what modeling component helps or hurts the generalization of OLN. We enumerate over different objectness cues, \ie, classification, centerness, IoU and Dice score, and their single-stage and two-stage configurations. The ablation is in \tableref{tab:configuration} and \tableref{tab:classifier_hurts}. 

\paragraph{{How much does the second stage help?}}\quad 
We notice that the second refinement head of two-stage models lead to better performance throughout all choices of objectness measure (\tableref{tab:configuration} - $a$ \textit{vs} $d$-$e$,  $b$ \textit{vs} $f$-$g$, and $c$ \textit{vs} $h$-$j$). This can be attributed to better bounding box regression which has additional layers following the first stage, as also noted by He et al.~\cite{he2017mask}. This trend can be also seen between other single-stage \textit{vs} multi-stage methods in \tableref{tab:sota}.

\paragraph{{Impact of different objectness cues.}}\quad We perform ablations on what objectness cue hurts or helps each stage of OLN. We start with single-stage models ($a$-$c$) where the box regression pipeline is analogous to that of standard RPN except the transformed box coordinates. We compare classification, centerness and IoU scores in \tableref{tab:configuration}; both localization-based scores outperform classification, and the centerness shows the best AR. Among the two-stage configurations, model-($d$) is comparable to Faster R-CNN, where only the last class score is used at testing. Again, we observe the overall superiority of localization-based objectness learning. We also notice that the second stage prefers IoU learning over centerness learning, whichever one of them is used in the first stage ($f$ \textit{vs} $g$ and $h$ \textit{vs} $i$). Intuitively, IoU measure is sensitive to both location and shape of the detection, and thus better captures the quality of variable-sized boxes in the second stage. On the other hand, centerness can be a more suitable measure for the first region proposal stage, where the shape of the anchors are fixed. Overall, the combination of centerness-\textit{then}-IoU ($i$) shows the best performance, implying their own contributions to the generalization. We also demonstrate that a different choice of localization quality also performs well ($j$). For the rest of the paper, OLN-RPN and OLN-Box refer to model-($c$) and ($i$) respectively.

\begin{table}[]
\centering
\resizebox{\linewidth}{!}{
\begin{tabular}{lll|cccc}
\hline
                         &          1st-stage & 2nd-stage &{\small{AR10}}  & \small{AR30}  & \small{AR100}  & \small{AR300} \\
\hline \hline

& \multicolumn{2}{l|}{\textit{\textbf{Single-stage:}}}   & & & &  \\
a. & Class              &     -     & 7.8  & 12.8  & 20.0  & 27.9  \\
b. & IoU                &     -     & 9.7  & 15.8  & 22.2  & 28.3    \\
c. & \textbf{Center}&   -    & \textbf{11.7}  & \textbf{19.4}  & \textbf{27.4}  & \textbf{33.7}  \\
\hline 
& \multicolumn{2}{l|}{\textit{\textbf{Two-stage:}}} & & & & \\
d. & Class~\tablefootnote{
For model-($d$), classification score from the first stage (RPN) is not used at inference; same as in the Faster R-CNN.}            & Class        & 11.9  & 17.6  & 25.5  & 34.4  \\
e. & Class             & Class        & 12.0  & 17.8  & 25.6  & 34.3   \\
f. & IoU               & Center       & 16.1  & 24.7  & 32.1  & 37.3  \\
g. & IoU               & IoU          & 16.2  & 25.0  & 32.3  & 37.5   \\
h. & Center            & Center       & 17.6  & 25.8  & 32.7  & 38.1   \\
i. & \textbf{Center}   & \textbf{IoU} & \textbf{17.7}  & \textbf{26.0}  & \textbf{32.7}  & \textbf{38.1}      \\
j. & Center            & Dice         & 17.2  & 25.3  & 32.2  & 37.5   \\
\hline
\end{tabular}}
\caption{\small{What objectness cues lead to better generalization? How much does the second stage help?} Different OLN configurations on {VOC} $\rightarrow$ {Non-VOC} setting.}
\label{tab:configuration}
\end{table}

\begin{table}[]
\centering
\resizebox{\linewidth}{!}{
\begin{tabular}{ll|cccc}
\hline
                              1st-stage & 2nd-stage &{\small{AR10}}  & \small{AR30}  & \small{AR100}  & \small{AR300} \\
\hline \hline
 \textbf{Center}&   -    & \textbf{11.7}  & \textbf{19.4}  & \textbf{27.4}  & \textbf{33.7}   \\
 Center + Class &   -    & 10.7  & 15.7  & 21.1  & 26.6   \\
 \hline
\textbf{Center}  & \textbf{IoU}  & \textbf{17.7}  & \textbf{26.0}  & \textbf{32.7}  & \textbf{38.1}  \\
 Center          & IoU + Class   & 13.4  & 19.4  & 27.3  & 35.2  \\
 Center + Class  & IoU           & 14.2  & 20.5  & 26.8  & 33.9  \\
 Center + Class  & IoU + Class   & 13.3  & 18.9  & 26.5  & 33.8  \\
 Center          & Class         & 13.0  & 18.7  & 26.5  & 34.6  \\
 Class           & Class         & 12.0  & 17.8  & 25.6  & 34.3 \\
\hline
\end{tabular}}
\caption{\small{How much does classifier hurt?} Different OLN configurations on {VOC} $\rightarrow$ {Non-VOC} setting.}
\label{tab:classifier_hurts}
\end{table}

\paragraph{{How much does classifier hurt?}}\quad
\tableref{tab:classifier_hurts} shows the results by adding a classifier branch to the best-performing OLN-RPN ($c$) and OLN-Box ($i$) models. Note that all the used objectness scores are geometrically averaged at test time. Throughout the single-stage and two-stage configurations, we observe a consistent drop in AR when adding a binary classifier. Largest drops are seen when having the classifiers in both stages. This validates our hypothesis that discriminative learning of object-or-not classification impedes generalization of object proposals, and that a pure localization-based objectness is the key to generalization.

\begin{figure}
\centering
\includegraphics[width=\linewidth]{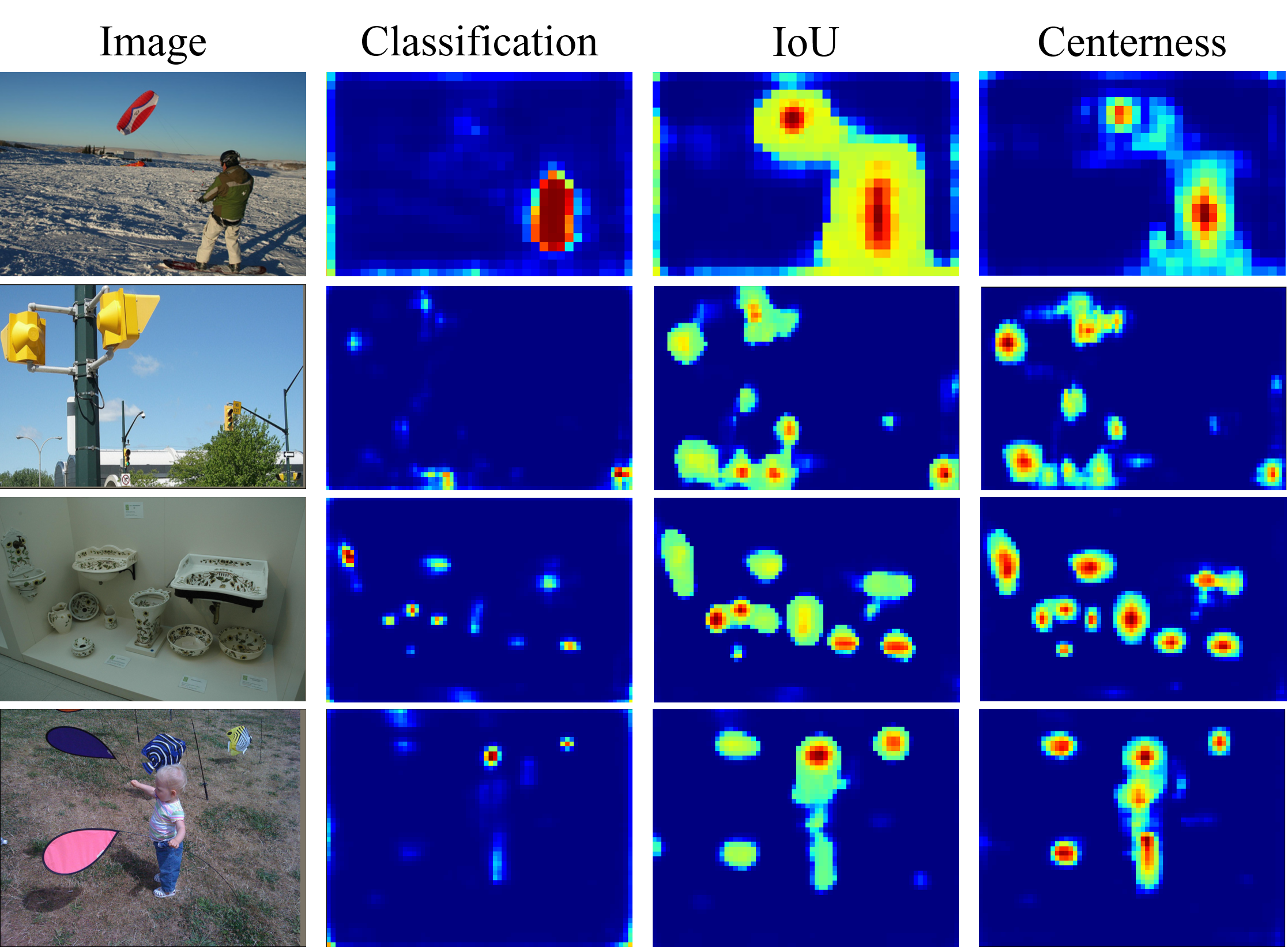}
\vspace{-4mm}
\caption{\small{Visualization of score maps of classification, IoU and centerness, obtained from OLN-RPN stage.}}
\label{fig:heatmap}
\end{figure}

\begin{table}[]
\centering
\resizebox{\linewidth}{!}{
\begin{tabular}{l|c|ccc|cc}
\hline
& B.G.  & RPN          & RPN       & RoI    &   &  \\
& sample & upper       & lower      & head       & AR10  &  AR100\\
& ratio      & thr.          & thr.      & thr. & & \\
\hline \hline
\multirow{2}{*}{\textbf{OLN-Box}}    & 0/256 & 0.3 & 0.1 &  0.3  & \textbf{17.7}     & \textbf{32.7} \\
                & \textbf{1/256} & 0.3 & 0.1 &  0.3  & \textbf{17.7}     & {32.6} \\
\hline
\multirow{4}{*}{\textbf{FR-CNN}}      & 128/256~\tablefootnote{Default anchor sampling ratio of the Faster R-CNN.} & {0.7} & {0.3} &  {{0.5}}  & \underline{{11.7}}     & \underline{{25.5}} \\
  & 128/256 & 0.3  & 0.1  & 0.3  & {7.3}  & {17.8} \\
  & 1/256 &{0.7} & {0.3} &  {{0.5}}  & {{10.1}} & {{21.8}} \\
  & 1/256   & 0.3  & 0.1  & 0.3  & {6.7}  & {14.9} \\
 \hline
  & 128/256 & {0.7} & {0.3} &  {{0.5}}  & \underline{{13.1}}     & \underline{{28.4}} \\
 \textbf{FR-CNN} & 128/256 & 0.3  & 0.1  & 0.3  & {8.3}  & {18.8} \\
 \textbf{*filtered*} & 1/256 & {0.7} & {0.3} & {{0.5}}  & {{12.0}} & {{25.8}} \\
  & 1/256   & 0.3  & 0.1  & 0.3  & {7.9}  & {16.5} \\
 
\hline
\end{tabular}
}
\caption{\small{Controlling B.G. sampling ratio and pos/neg definition.}}
\label{tab:sampling}
\end{table} 

\begin{table}[]
\centering
\resizebox{0.9\linewidth}{!}{
\begin{tabular}{l|c|c|c}
\hline
            & mask IoU      & box-AR10 & mask-AR10 \\ 
\hline \hline
Mask-RCNN~\cite{he2017mask} & & 11.8     & 9.1  \\
\hline
OLN-Box     & N/A           & 17.7      & -    \\
OLN-Mask    &               & 17.9       & 15.8 \\
OLN-Mask    & \checkmark    & 18.3       & 16.9 \\
\hline
\end{tabular}}
\caption{\small{Does mask prediction learning help?}}
\label{tab:mask_head}
\end{table}

\paragraph{{Impact of positive / negative sampling ratio.}}\quad
Is the gain of \OURS{} from the fact that we sample around the objects and avoid penalizing the unlabeled background objects? or does the permissive definition for positive helps generalization? In \tableref{tab:sampling} we run Faster R-CNN with low sampling of background anchors similar to OLN. We validate that changing the sampling ratio does not help the baseline. This is because a balanced ratio (\eg, 1:1) of explicit negative vs positive samples is required for the binary classifier. \tableref{tab:sampling} also shows various definition of positive / negative samples and that OLN-Box still outperforms Faster R-CNN and *filtered* models by a healthy margin. We think it is becuase OLN can learn more from the positives using localization cues instead of binary classification.

\paragraph{{Does mask prediction learning help?}}\quad
In \tableref{tab:mask_head} we show the results for the mask-extended OLN model, OLN-Mask. We validate that incorporation of additional localization quality learning, \ie, mask-IoU, can further improve both the box-AR and mask-AR performances.

\paragraph{{Visualization of VOC to Non-VOC transfer on COCO.}}\quad
We visualize class-agnostic Mask R-CNN and our \OURS{}-Mask trained on VOC categories on COCO in \figureref{fig:coco}. The images come from COCO validation set. We observe Mask R-CNN tends to miss out-of-VOC class objects even when they appear in the dominant and salient region in an image, \eg, \textit{refrigerator} (left column) and \textit{giraffes} (right column). We believe this is because classification leads to learning just the union of 20 VOC categories and use that knowledge to detect objectness. On the other hand, \OURS{} can generalize well to detect many novel objects, such as \textit{fridge, microwave, vases, and giraffes}, showing that localization leads to learning a more general notion of objectness.

\begin{figure*}[t]
\begin{center}
\resizebox{\linewidth}{!}{
\begin{tabular}{@{}c@{\hskip 0.002\linewidth}c@{\hskip 0.002\linewidth}c@{\hskip 0.002\linewidth}c@{}}
\includegraphics[width=0.198\linewidth]{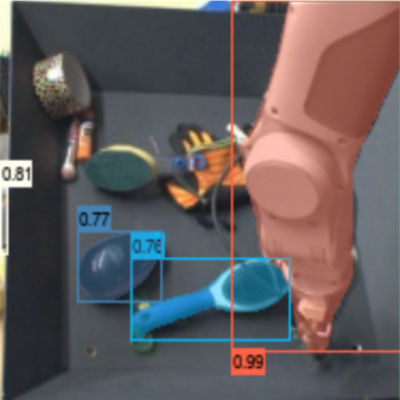} &
\includegraphics[width=0.348\linewidth]{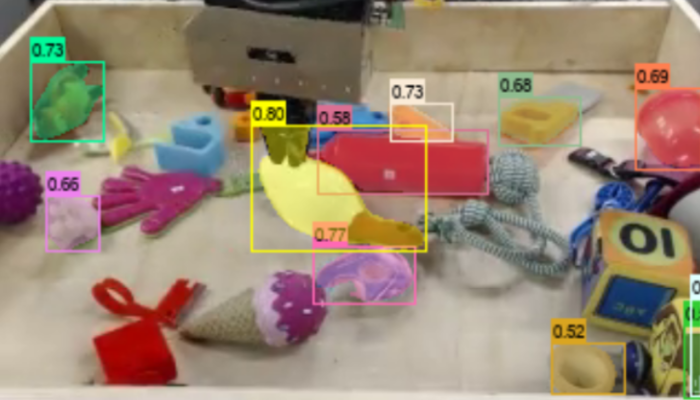} &
\includegraphics[width=0.248\linewidth]{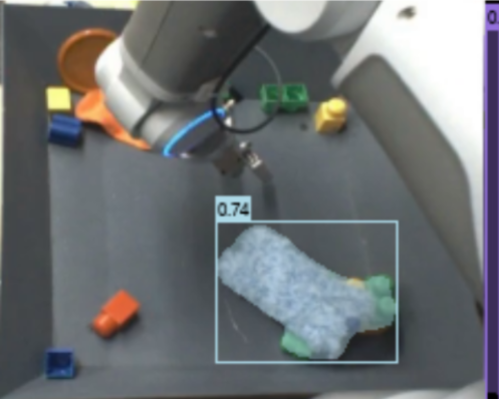} &
\includegraphics[width=0.198\linewidth]{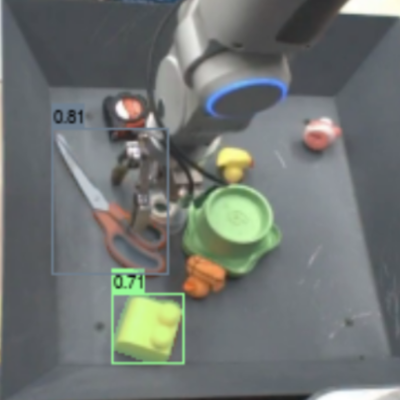}
 \\
\includegraphics[width=0.198\linewidth]{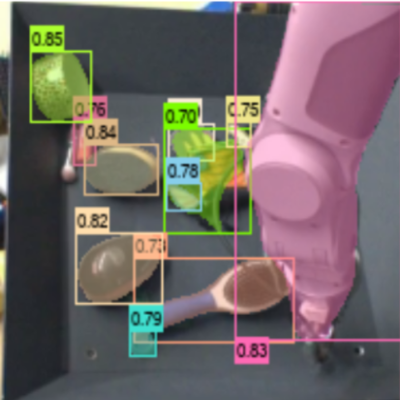} &
\includegraphics[width=0.348\linewidth]{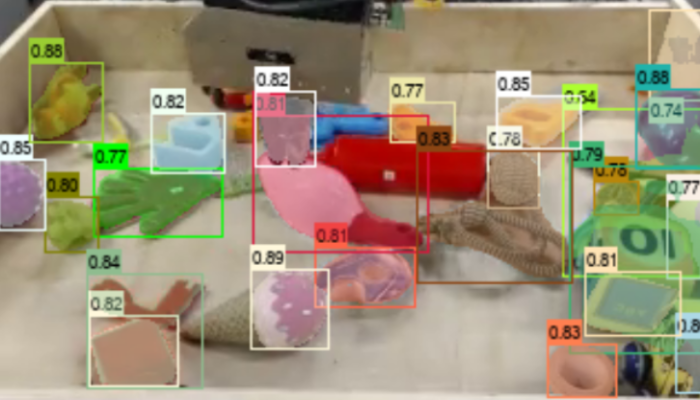} &
\includegraphics[width=0.248\linewidth]{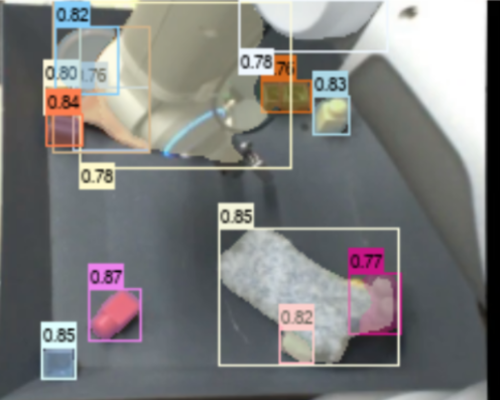} &
\includegraphics[width=0.198\linewidth]{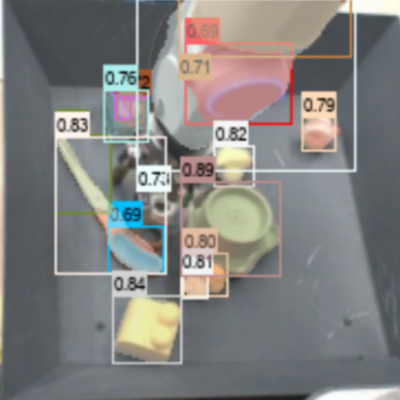} 
\\
\end{tabular}}
\end{center}
\vspace{-4mm}
\caption{\small{Visualization of {COCO} $\rightarrow$ {RoboNet} generalization. (Top: Mask R-CNN, Bottom: OLN-Mask.)}}
\label{fig:robo}
\vspace{0mm}
\end{figure*}
\begin{figure*}[t]
\begin{center}
\resizebox{\linewidth}{!}{
    \begin{tabular}{@{}c@{\hskip 0.002\linewidth}c@{\hskip 0.015\linewidth}c@{\hskip 0.002\linewidth}c@{}}
\includegraphics[width=0.225\linewidth]{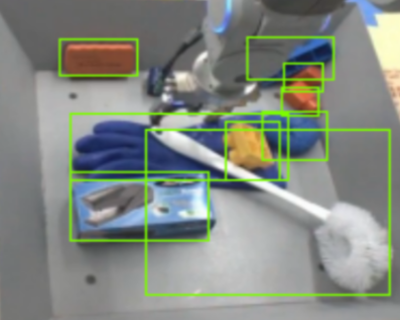} &
\includegraphics[width=0.225\linewidth]{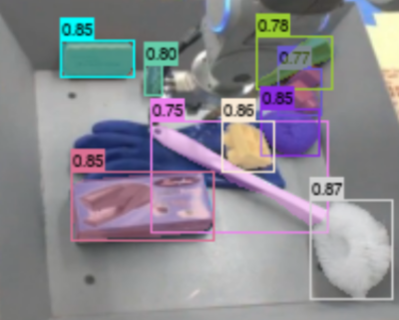} &
\includegraphics[width=0.271\linewidth]{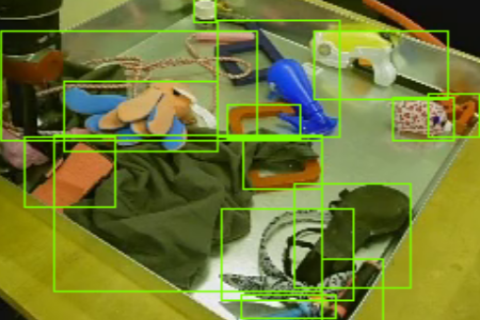} &
\includegraphics[width=0.271\linewidth]{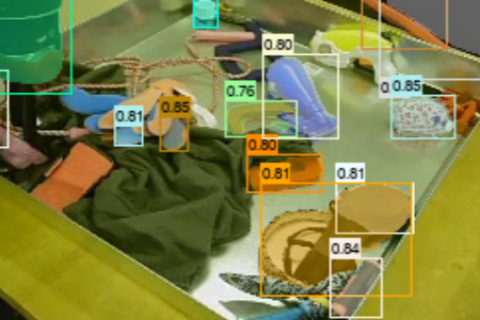}  \\
\small{(a) GT boxes} & \small{(b) \OURS{} prediction} & \small{(c) GT boxes} & \small{(d) \OURS{} prediction}
\end{tabular}}
\end{center}
\vspace{-4mm}
\caption{\small{Analysis of failure modes of OLN on {COCO} $\rightarrow$ {RoboNet} generalization. For each sample, we show the ground-truth boxes on the left and our \OURS{}-Mask predictions on the right.}}
\label{fig:robo_fail}
\end{figure*}

\paragraph{{Visualization of the learnt objectness map.}}\quad
In \figureref{fig:heatmap}, we visualize the confidence heatmap of different localization cues \textit{vs} classifier. We can see the IoU and centerness cues generalize better to novel objects than classification. For example, both IoU and centerness capture the flying kite, the traffic lights, handicrafts, the child and his/her toys, while the classifier head misses them. We use model-$a$, $b$, and $c$ in \tableref{tab:configuration} for classification, IoU and centerness visualization respectively.

\subsection{Open-World Class-Agnostic Detection}
\label{sec:robonet_detection}
Our ultimate goal is to learn a generalizable object detector that can detect \textit{every} object in our open world. Taking a step towards this goal, we investigate the \textit{open-world, class-agnostic detection} ability on highly diverse and hard-to-name objects in RoboNet dataset. We train a model on COCO and test on held-out RoboNet~\cite{dasari2019robonet} dataset.

\paragraph{Rigorous evaluation.} \quad
Chavali et al.~\cite{chavali2016object} report the vulnerability of current evaluation protocol for object proposals when evaluating on a \textit{partially-annotated} dataset such as COCO, because it fails to capture the performance (\eg, reward the recall) on all the non-COCO categories that are not annotated in the ground truth. To handle this issue, we \textbf{exhaustively annotated bounding boxes for the presence of \textit{all} objects in RoboNet images}. This allows us to measure \textit{\textbf{Average Precision}} as in standard object detection evaluation. Such evaluation captures whether a model can generate \textit{a few but highly-precise} detections. We use NMS threshold of 0.5 and evaluate box-AP on top 100 proposals in this experiment.

\paragraph{Exhaustive box annotations.} \quad
To enable the evaluation of our task also by the standard AP metric, we carefully annotated the RoboNet dataset for the presence of all objects in an exhaustive fashion. RoboNet is a large-scale robot manipulation video dataset collected for pre-training reinforcement learning~\cite{dasari2019robonet}. The dataset contains very challenging clutter and object diversity that we find suitable for evaluating general objectness learning. Since the video data contains lots of redundancy at frame level, we resort to random sampling to construct a concise yet diverse evaluation set. Concretely, we randomly sampled 109 videos from the 162417 videos in the training set (roughly 1/1500) and for each video we select only the center frame. The parameters are picked by visually inspecting the diversity of resulting data (See \figureref{fig:robo_fail}-(a, c) for examples). Our annotated dataset consists of total 109 frames, 1277 objects, and 34.8\% of small objects, which are those with long side $<$ 32 pixels (same as COCO). For object bounding box annotation, we adopt the method of labeling extreme points as proposed in~\cite{kuznetsova2018open}, which results in better accuracy and efficiency for the raters.

To exclude the background scene in the image which could contain hard-to-define objects/building structures, we label the object bin as well and crop out only the within-bin region for evaluation. This removes most of the ambiguity in objectness evaluation and allows us to have reasonably exhaustive annotation of objects for detection evaluation via average precision. To validate whether the size is large enough, we trained 5 independent runs of our model and observe a variance of 0.17 AP, showing that the results are quite reproducible on our dataset size.

\begin{table}[]
\centering
\resizebox{\linewidth}{!}{
\begin{tabular}{l|cccccc}
\hline
                                & AP    & AP\textsubscript{50}  & AP\textsubscript{75} & AP\textsubscript{S} & AP\textsubscript{M}  & AP\textsubscript{L} \\
\hline \hline
RPN~\cite{fasterNIPS2015}       & 13.7 & 28.1 & 12.2 & 15.6 & 22.8 & 10.3 \\
Faster R-CNN~\cite{fasterNIPS2015} & 10.7 & 20.9 & 10.0  & 7.6  & 18.0 & 8.7  \\
Mask R-CNN~\cite{he2017mask}      & 11.4 & 22.3 & 10.4 & 8.7  & 18.6 &  7.9 \\
\hline
\textbf{\OURS{}-Box (Ours)}     & \textbf{24.6} & \textbf{47.1} & \textbf{23.0} & \textbf{20.9} & \textbf{32.1} & \textbf{15.4} \\
\textbf{\OURS{}-Mask (Ours)}    & \textbf{26.4} & \textbf{50.1} & \textbf{24.7} & \textbf{22.9} & \textbf{34.0}  & \textbf{15.6}\\
\hline
\end{tabular}}
\caption{\small{AP results of open-world class-agnostic detection in {COCO} $\rightarrow$ {RoboNet} generalization.}}
\vspace{-2mm}
\label{tab:coco_robonet}
\end{table}

\paragraph{Results.} \quad
In this \textbf{rigorous evaluation setting, larger gains are revealed for OLN models} than in other experiments. In \tableref{tab:coco_robonet} we compare {COCO} \textrightarrow {RoboNet} generalization performance of OLN-Box and OLN-Mask models and class-agnostic Faster R-CNN and Mask R-CNN baselines. \OURS{} models greatly outperform RPN and Faster R-CNN by \textbf{+12.7} and \textbf{15.7} points AP, respectively. We also observe that the improvement in AP\textsubscript{IoU=.50} is significantly larger, by \textbf{+22.0} and \textbf{29.1} points AP. These results provide strong evidence that \OURS{} can achieve both \textit{high precision and recall} in detecting novel objects, and that \textit{a few generated OLN proposals ($\sim$100) can be directly used as final bounding boxes} for all objects in an image, \ie, class-agnostic detection.

\paragraph{Visualization and failure cases.} \quad
\figureref{fig:robo} visualizes the box and mask predictions from baseline class-agnostic Mask R-CNN \textit{vs} our \OURS{}-Mask model. \OURS{} is able to detect many novel objects (\eg gloves, toys, tape, parts of toys), while the baseline misses most of them. \figureref{fig:robo_fail} presents an analysis of \OURS{} failure modes by comparison with the ground-truth boxes. The false positive example is where \OURS{} detects \textit{part} of an object, \eg, it detects \textit{handle} and \textit{head} of a cleaning brush as two individual objects. 

\subsection{More cross-dataset generalization}
\label{sec:cross_dataset}
We further study the generalization ability of OLN from one data source to different data sources.

\subsubsection{Generalization to Objects365}
Objects365 is a large-scale object detection dataset consisting of 365 object categories from our daily lives, which are a super-set of 80 COCO categories. We train a model on COCO and test on non-overlapping classes on this dataset. We follow the same AR@k evaluation protocol in \sectionref{sec:cross_category}, treating all the 365 classes as a single `object' class, and leaving out the detection boxes on \textit{seen} COCO 80 classes when evaluating top-k recall.

\paragraph{Results.} \quad
\tableref{tab:coco_objects365} shows the {COCO} \textrightarrow {Objects365} generalization results. \OURS{} outperforms all the classification-based baselines for all number of proposals considered. Especially, the gains over the RPN and Faster R-CNN start from +4.5 and 3.3 AR10 and achieve large gaps of +8.2 and 7.6 points in AR100. 

\begin{table}[]
\centering
\resizebox{0.86\linewidth}{!}{
\begin{tabular}{l|cccc}
\hline
                        & \small{AR10}   & \small{AR20}  & \small{AR50}  & \small{AR100} \\
\hline \hline
RPN~\cite{fasterNIPS2015}            & 8.8     & 12.5  & 18.1  & 22.5  \\
Faster R-CNN~\cite{fasterNIPS2015}     & 10.1    & 13.7  & 19.1  & 23.1  \\
Mask R-CNN~\cite{he2017mask}        & 10.1    & 13.7  & 19.2  & 23.3  \\
\hline
\textbf{\OURS{}-Box (Ours)}   & \textbf{13.1}    & \textbf{18.8}  & \textbf{26.0}  & \textbf{30.6}  \\
\textbf{\OURS{}-Mask (Ours)}  & \textbf{13.3}    & \textbf{18.9}  & \textbf{26.0}  & \textbf{30.7}  \\
\hline
\end{tabular}
}
\caption{\small{AR results of {COCO} $\rightarrow$ 
{Objects365} generalization.}}
\vspace{-2mm}
\label{tab:coco_objects365}
\end{table}

\begin{figure*}[t]
\begin{center}
\resizebox{\linewidth}{!}{
\begin{tabular}{@{}c@{\hskip 0.002\linewidth}c@{\hskip 0.002\linewidth}c@{}}
\includegraphics[width=0.331\linewidth]{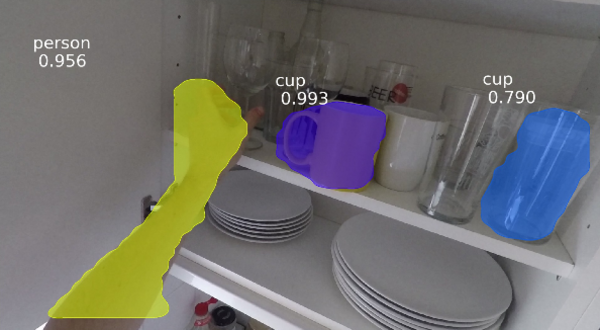} &
\includegraphics[width=0.331\linewidth]{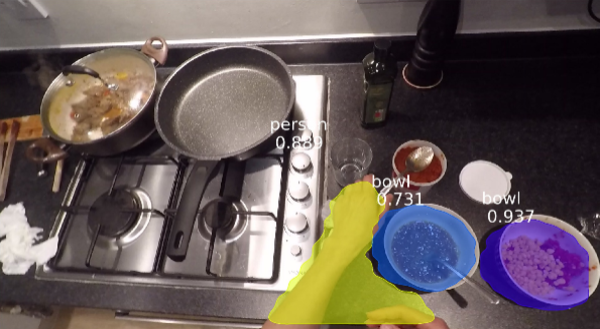} &
\includegraphics[width=0.331\linewidth]{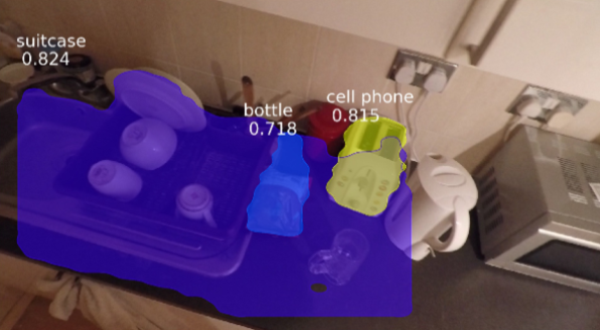} \\
\includegraphics[width=0.331\linewidth]{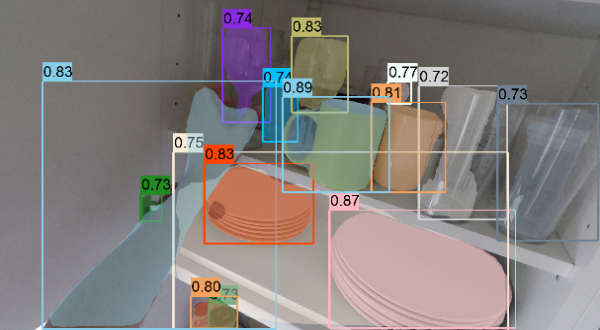} &
\includegraphics[width=0.331\linewidth]{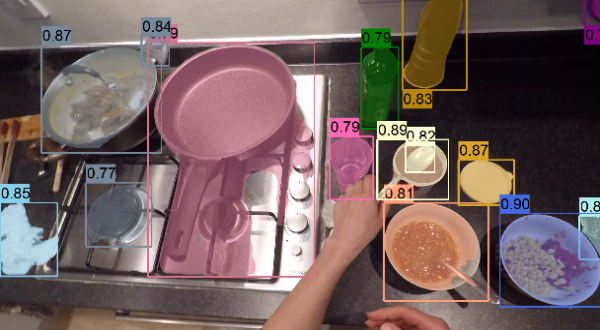} &
\includegraphics[width=0.331\linewidth]{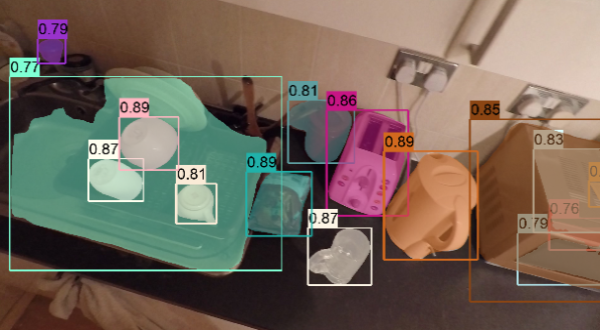} \\
\end{tabular}}
\vspace{-4mm}
\end{center}
\caption{\small{Visualization of \textit{COCO} $\rightarrow$ \textit{EpicKitchens} generalization. We compare the EpicKitchen's official annotations~\cite{Damen2020Collection} obtained from Mask R-CNN (left) vs. ours from OLN-Mask (right).}}
\vspace{1mm}
\label{fig:epic}
\end{figure*}

\begin{table*}[]
\centering
\resizebox{\textwidth}{!}{
\begin{tabular}{l|l|cccc|cccc}
\hline
{Detector} & {Proposals}                & AR    & AR\footnotesize{ rare} & AR\footnotesize{ comm.}  & AR\footnotesize{ freq.}
                                & AP    & AP\footnotesize{ rare} & AP\footnotesize{ comm.}  & AP\footnotesize{ freq.}\\
\hline \hline
\multirow{2}{*}{{FR-CNN}}
& RPN ~\cite{fasterNIPS2015}
& 32.3 {\color[HTML]{FFFFFF} --------} 
& 14.6 {\color[HTML]{FFFFFF} --------} 
& 30.2 {\color[HTML]{FFFFFF} --------}  
& 41.9 {\color[HTML]{FFFFFF} -------}
& 24.1 {\color[HTML]{FFFFFF} --------} 
& 13.0 {\color[HTML]{FFFFFF} --------} 
& 23.7 {\color[HTML]{FFFFFF} --------}  
& 28.9 {\color[HTML]{FFFFFF} --------} \\

& OLN-RPN
& \textbf{33.8} {\color[HTML]{32CB00} \textbf{(+1.5)}}
& \textbf{19.9} {\color[HTML]{32CB00} \textbf{(+5.3)}}  
& \textbf{32.1} {\color[HTML]{32CB00} \textbf{(+1.9)}}
& \textbf{41.4} {\color[HTML]{32CB00} \textbf{(-0.5)}}
& \textbf{25.5} {\color[HTML]{32CB00} \textbf{(+1.4)}}
& \textbf{16.4} {\color[HTML]{32CB00} \textbf{(+3.4)} } 
& \textbf{25.4} {\color[HTML]{32CB00} \textbf{(+1.8)}}
& \textbf{29.3} {\color[HTML]{32CB00} \textbf{(+0.4)}}\\

\hline
\end{tabular}}
\vspace{1mm}
\caption{\small{Long-tail detection on LVIS v0.5 using Faster R-CNN~\cite{fasterNIPS2015}.}}
\label{tab:lvis}
\end{table*}

\subsubsection{Generalization to EpicKitchens}
EpicKitchens is the largest egocentric vision dataset containing the wearers' different kitchen activities like washing and cooking, with many different objects like food, kitchenware and appliances. It provides the automatic mask annotations of 66M objects, which they extracted by using the off-the-shelf Mask R-CNN~\cite{he2017mask} trained on COCO dataset. Therefore, we use their official mask annotations as our baseline in this experiment.

\paragraph{Results.} \quad 
AR or AP are difficult to compute on EpicKitchens because there are no \textit{groundtruth} boxes or masks. Therefore, we visually compare the EpicKitchens mask annotations and our \OURS{}-Mask prediction outputs in \figureref{fig:teaser} and \ref{fig:epic}, which are randomly selected examples. We observe that \OURS{}-Mask is able to detect almost complete set of objects in an image, while the baseline Mask R-CNN misses most out-of-sample objects. To illustrate, \OURS{} is able to detect \textit{a stack of dishes, dough balls, a toaster, half side of a microwave and its front window}, as shown in \figureref{fig:teaser} and \ref{fig:epic}. These are novel categories outside of the COCO vocabulary, and appear in different configurations and viewpoints rarely seen in COCO. This demonstrate the clear benefits of learning localization for novel object detection.

\begin{figure}
\centering
\includegraphics[width=\linewidth]{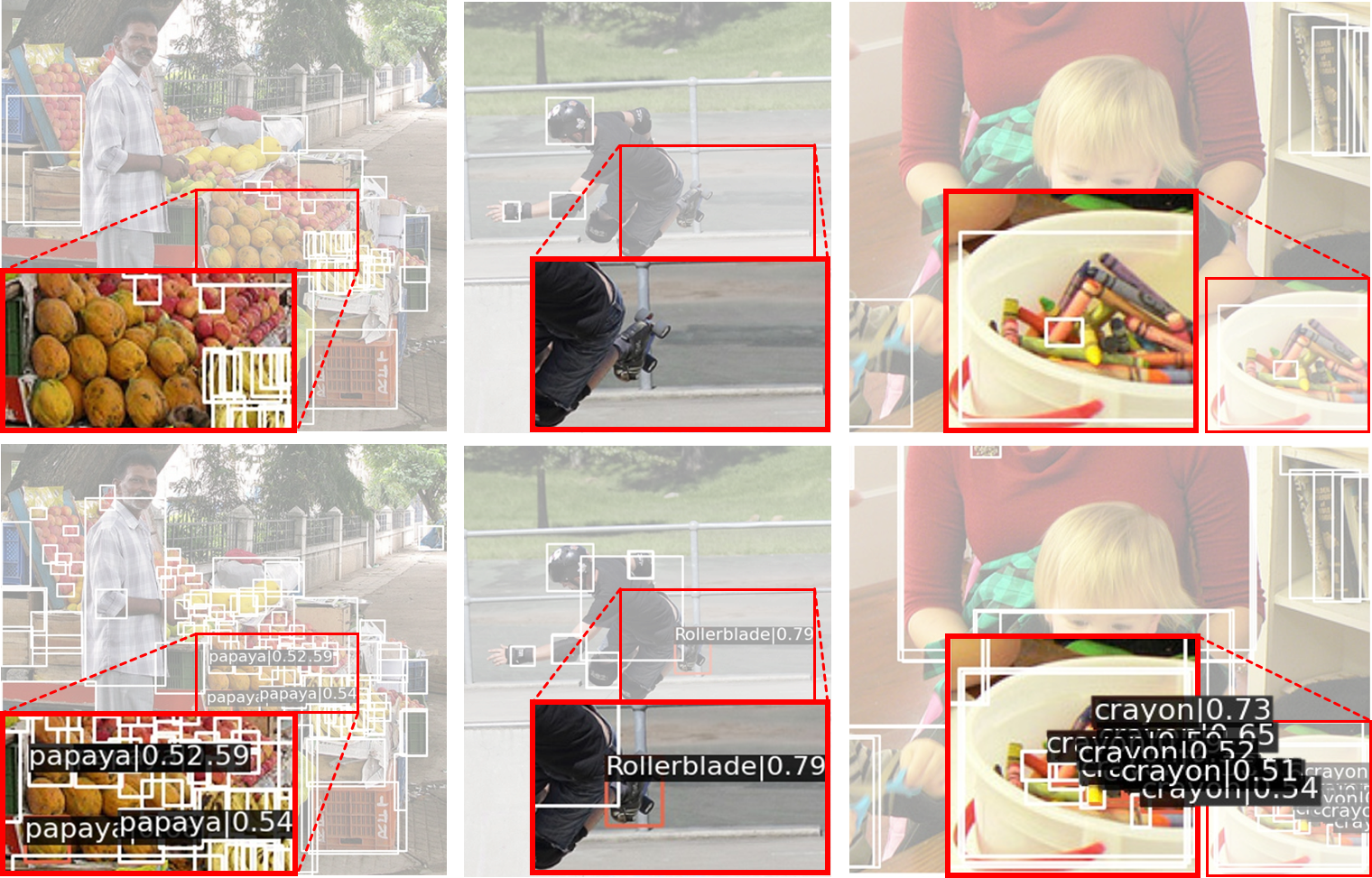}
\caption{{Visualization of long-tail detection on LVIS using Faster R-CNN. (Top: RPN, Bottom: OLN-RPN.)}}
\label{fig:lvis}
\end{figure}

\subsection{Impact on long-tail object detection}
\label{sec:application}

Large vocabulary detection has gained much attention lately in research community \cite{gupta2019lvis,wang2020devil,ghiasi2020simple,tan20201st} because the problem captures the Zipf's law of visual world, i.e. most categories appear with low frequency. We want to see whether \OURS{} can aid downstream detection in a challenging long-tail setting, on LVIS dataset~\cite{gupta2019lvis}, \eg, only 1-10 training samples are available for \textit{rare} categories. \tableref{tab:lvis} shows that a drop-in replacement of RPN\cite{fasterNIPS2015} with \OURS{}-RPN helps the overall average recall (AR) by +1.5, where most gains come from rare (\textbf{+5.3}) and common categories (+1.9). The 5-point gain in rare categories recall  demonstrates that OLN can benefit the long-tail of large vocabulary detection. In fact, this gain transfers quite well to overall average precision (AP) improvement +1.4, where most gains come from the rare (\textbf{+3.4}) and common (+1.8) categories. We choose $k=300$ detections per image, following the convention of the community. Our baseline matches the performance of Faster R-CNN reported by \cite{wang2020frustratingly} in the same setting. 

Figure \ref{fig:lvis} presents a few challenging cases of long-tail detection on \textit{rare} categories - piles of papaya, crayon, and a small roller-blade. Compared to the baseline, a drop-in OLN-RPN replacement enables better detection on these challenging cases. We can see many missed papayas, crayons and the small roller-blade are now correctly localized and classified.

\section{Conclusion}
In this paper we tackle the challenging problem of learning novel object proposals. Observing the tendency of existing proposals to overfit to training categories, we propose a simple yet effective framework (\OURS{}) that learns to propose novel objects by learning localization cues (centerness, IoU, and regression) instead of binary classification. Experiments show that \OURS{} outperforms existing methods on cross-category generalization on COCO, as well as cross-dataset settings on RoboNet, Object365, and EpicKitchens. Moreover, a drop-in replacement of \OURS{} improves the performance on large vocabulary and egocentric video object detection on LVIS dataset. We believe \OURS{} is a step forward in open-world novel object understanding with many applications.

{\small
\bibliographystyle{ieee_fullname}
\bibliography{egbib.bbl}
}

\end{document}